# An attempt to generate new bridge types from latent space of generative adversarial network

Hongjun Zhang

Wanshi Antecedence Digital Intelligence Traffic Technology Co., Ltd, Nanjing, 210016, China

583304953@QQ.com

**Abstract:** Try to generate new bridge types using generative artificial intelligence technology. Symmetric structured image dataset of three-span beam bridge, arch bridge, cable-stayed bridge and suspension bridge are used. Based on Python programming language, TensorFlow and Keras deep learning platform framework, as well as Wasserstein loss function and Lipschitz constraints, generative adversarial network is constructed and trained. From the obtained low dimensional bridge-type latent space sampling, new bridge types with asymmetric structures can be generated. Generative adversarial network can create new bridge types by organically combining different structural components on the basis of human original bridge types. It has a certain degree of human original ability. Generative artificial intelligence technology can open up imagination space and inspire humanity.

## 0　Introduction

Bridge designers cannot draw wild and imaginative patterns like artists. Bridge design must first meet the structural force requirements. Unreasonable structural design will greatly increase the cost of bridge construction, while mechanically infeasible solutions cannot be implemented. Although bridge type innovation is strictly constrained by structural mechanics, it is not entirely impossible today because the history of large-scale bridge construction by humans is only over 100 years, and there is still much unknown space to explore in the field of bridges. In the past decade, the progress of generative artificial intelligence technology has been astonishing, providing new means for bridge innovation.

The author's previous paper[1] used Variational Autoencoder (VAE) to generate several technically feasible new combination bridge types, but they were only a simple superposition of two bridge types in the dataset, and the innovative ideas were too single. This is because the variational autoencoder distributes samples with the same label in the same region, and there are overlapping areas between different regions. In order to meet the minimum binary cross entropy requirement between the input samples and the generated images, the sampling at the overlapping area always has features of different peripheral bridge types. This feature allows it to generate composite bridge types, but also limits its ability to freely create. (Note: Variational autoencoder does not lack originality. The reason why it can only be simply superposition here is related to the pre-conditions such as the dataset and model parameters.)

Generative Adversarial Networks (GANs) have achieved great success in fields such as computer vision and natural language processing. They do not require pixel to pixel matching between input samples and generated images, but rather determine whether the visual statistical features of the images match. This makes them more creative than variational autoencoders. This article adopts generative adversarial network and attempts bridge innovation again based on the same dataset as before [1] (open source address of this article's dataset and source code: https://github.com/QQ583304953/Bridge-GAN).

## 1　Introduction to generative adversarial network

### 1.1 Overview

Generative artificial intelligence technology can be roughly divided into six categories:

Variational Autoencoders, Generative Adversarial Networks, Autoregressive Models and Transformers, Normalizing Flow Models, Energy Based Models and Diffusion Models, Multimodal Models[2]. Many categories do not have absolute boundaries and often merge and permeate with each other. The timeline for the development of contemporary generative artificial intelligence is shown in the following figure (Figure 1):

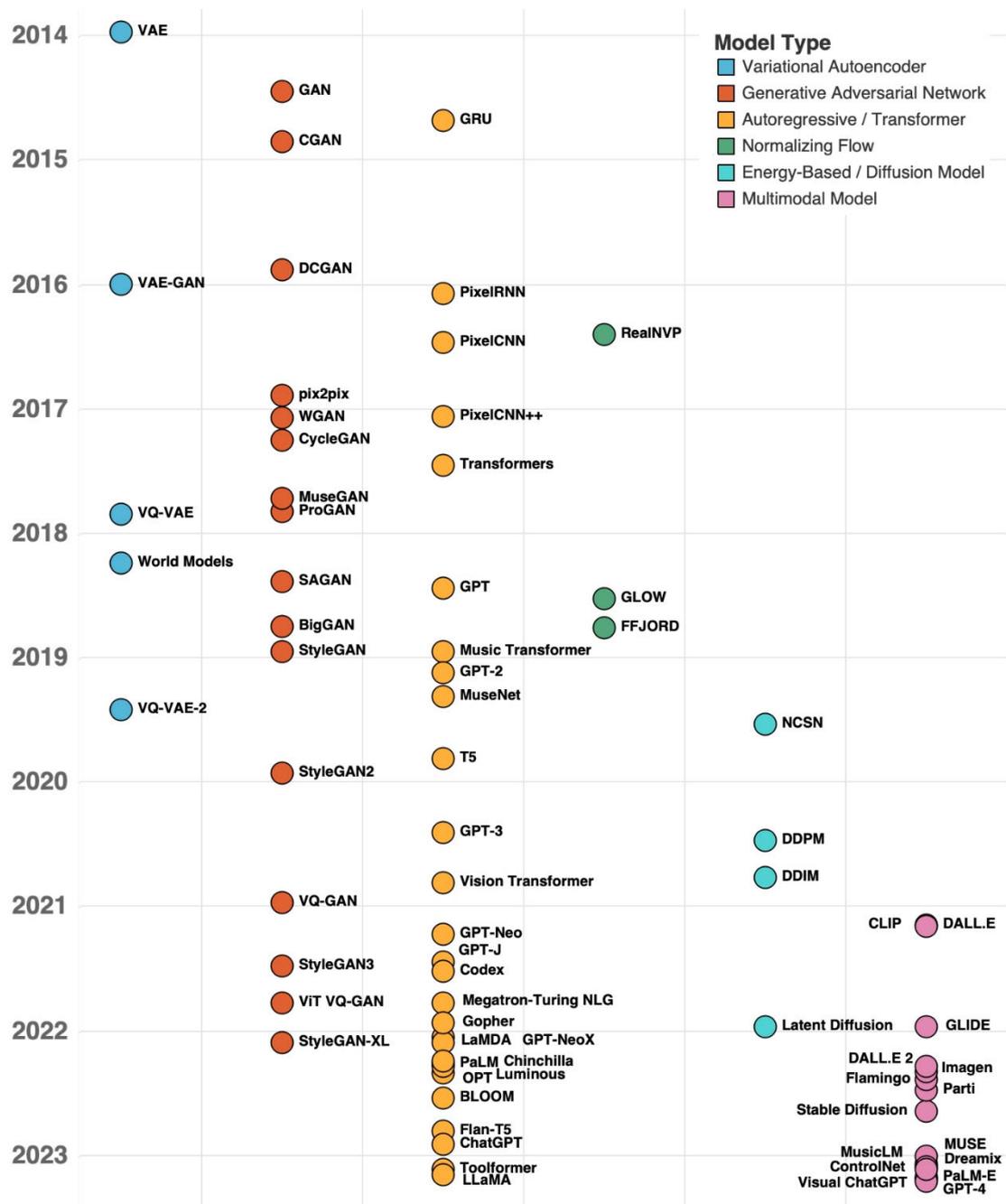

Fig.1 The timeline for the development of generative artificial intelligence

Data source: https://github.com/davidADSP/Generative_Deep_Learning_2nd_Edition/tree/main/docs/timeline.png

## 1.2 Generative adversarial network

Generative adversarial networks can replace variational autoencoders to learn the latent space of images. They can make the generated images almost indistinguishable from the training images statistically, thus generating realistic new images.

It includes a discriminator and a generator [3]. The function of the discriminator is to distinguish whether the input images are from the dataset (real) or the generator composite (fake) by learning the features of the input images (from the dataset and the generator composite). The function of the generator is to decode a random vector (points in the latent space) into a composite image with dataset features through learning under the guidance of the discriminator. Two networks are trained alternately, and their abilities are synchronously improved until the generated images by the generator can be astonishingly realistic, making it difficult for the discriminator to distinguish them. Please refer to the following figure (Figure 2) for details:

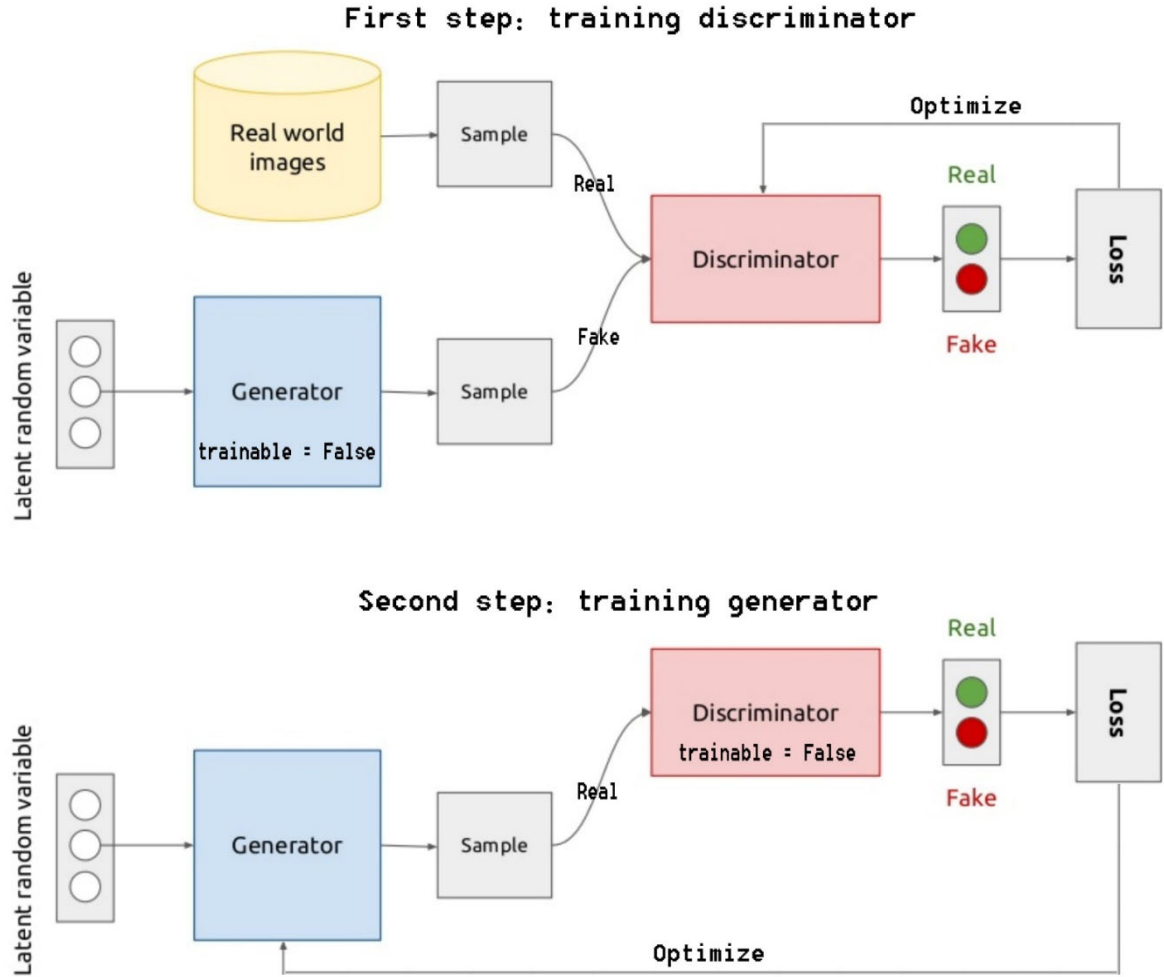

Fig.2 Architecture diagram of generative adversarial network

### 1.3 WGAN-GP

At the beginning of this practice, I attempted to use the standard GAN architecture, but encountered the problem of mode collapse. The generator could only output one type of bridge (such as V-shaped pier rigid frame beam bridge), so I switched to the WGAN-GP architecture[4].

(1) The WGAN-GP architecture uses the Wasserstein loss function instead of binary cross entropy.

The Wasserstein loss function[5] is $-1/n*\Sigma(y*p)$.

In the formula: n is the number of samples; The value of label y is 1 (real, or artificially set as real), -1 (fake); P is the output value of the discriminator (real scalar).

The loss during training the discriminator is $-1/n * \Sigma \{D(x) - D[G(z)]\}$.

The loss during training the generator is $-1/n * \Sigma \{D[G(z)]\}$.

In the formula: D (x) is the output value of the discriminator for real image; Z is the latent space sampling point; G (z) is the output value of the generator (image); D [G (z)] is the output value of the discriminator for generating image.

In order to minimize the loss, the gradient descent algorithm forces the discriminator to adjust the weight parameters to achieve the following results: D (x) as large as possible and D [G (z)] as small as possible. Similarly, the result of generator optimization is to maximize D [G (z)] as much as possible.

From a mathematical perspective, the Wasserstein loss function here has neither upper nor lower bounds, and there is no minimum value for the loss (while the binary cross entropy loss of GAN has a minimum value).

The influence of the Wasserstein loss function can be understood through imagination: ① During discriminator training, a counterfeiter (generator) imitates the artworks of a famous artist, and the discriminator compares and evaluates the counterfeits with the real artworks, striving to improve identification skills, scoring the real works as positive and the fake works as negative, and polarizing the scores as much as possible. ② During generator training, in order to get the discriminator to give the highest possible positive score, counterfeiters must do their best to improve their imitation skills. It can be seen that logically WGAN and GAN are basically the same.

In an ideal state, when the game reaches equilibrium: the discriminator is difficult to distinguish between real and fake, D (x) is approximately equal to D [G (z)], and the discriminator's loss approaches zero from a large negative value; The generator's ability has also improved to an excellent state, and the generator loss is approaching zero.

(2) During the training process, for the input images of real and fake categories, it is required that the discriminator score is polarized, and the discriminator's output can be any value (-∞,+∞). Under the influence of gradient descent algorithm, the gradient of the discriminator function (the discriminator is assumed to be a higher-dimensional function y=f(x) that maps an image as a scalar) becomes very steep, thus achieving the goal of minimal loss (negative value). But the steep gradient leads to poor model stability.

Therefore, it is necessary to apply constraints to make the discriminator function a K-Lipschitz function. The method is to add a gradient penalty[6] to the loss function of the discriminator, which will impose a penalty when the gradient norm of the discriminator function deviates from the K value. At the end of the training, the gradient of the discriminator function will approach the set K value (K value is taken as 1 here).

Calculation example: If the gradient of the discriminator function is [3,4], the gradient penalty is [sqrt (3 ^ 2+4 ^ 2) -1] ^ 2=16, the relative coefficient is 10, and the discriminator Wasserstein loss is -260, then the total discriminator loss is=Wasserstein loss+relative coefficient * gradient penalty term=-260+10 * 16=-100.

## 2 An attempt to generate new bridge types from latent space of generative adversarial network

### 2.1 Dataset

Using the dataset from the author's previous paper [1], which includes two subcategories for each type of bridge (namely equal cross-section beam bridge, V-shaped pier rigid frame beam bridge, top-bearing arch bridge, bottom-bearing arch bridge, harp cable-stayed bridge, fan cable-stayed bridge, vertical_sling suspension bridge, and diagonal_sling suspension bridge), and all are three spans (beam bridge is 80+140+80m, while other bridge types are 67+166+67m), and are structurally symmetrical.

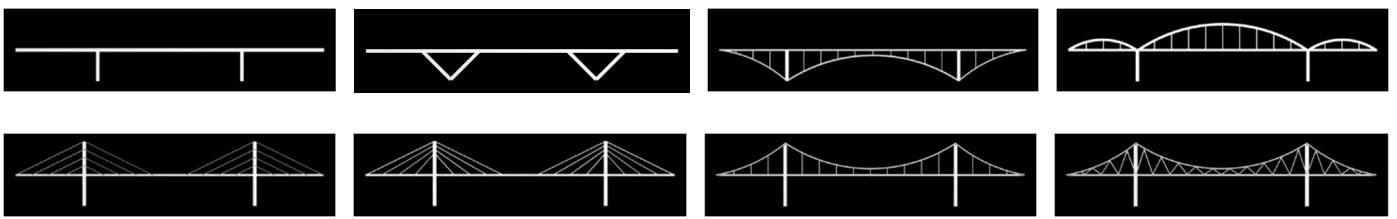

Fig.3 Grayscale image of each bridge facade

Each sub bridge type obtained 1200 different images, resulting in a total of 9600 images in the entire dataset.

### 2.2 Construction of generative adversarial network

Based on the Python3.8 programming language, TensorFlow2.6, and Keras2.6 deep learning platform framework, construct and train generative adversarial network. After testing, the 2-dimensional latent space can achieve relatively ideal results.

(1) Architecture of the critic (discriminator):

It receives input images from the dataset and synthesized by the generator, learning to correctly distinguish between the two.

First, create an input layer for the image, then passed to six Conv2D layers in sequence (activation function LeakyReLU), and finally flatten and connect it to the Dense layer (linear transformation). The output result is the scoring scalar value. The model summary is shown in the table below:

Tab.1 model summary of critic

| Layer (type) | Output Shape | Param # |
| --- | --- | --- |
| input_1 (InputLayer) | [(None, 128, 512, 1)] | 0 |
| conv2d (Conv2D) | (None, 64, 256, 64) | 1088 |
| leaky_re_lu (LeakyReLU) | (None, 64, 256, 64) | 0 |
| dropout (Dropout) | (None, 64, 256, 64) | 0 |
| conv2d_1 (Conv2D) | (None, 32, 128, 128) | 131200 |
| leaky_re_lu_1 (LeakyReLU) | (None, 32, 128, 128) | 0 |
| dropout_1 (Dropout) | (None, 32, 128, 128) | 0 |
| conv2d_2 (Conv2D) | (None, 16, 64, 128) | 262272 |
| leaky_re_lu_2 (LeakyReLU) | (None, 16, 64, 128) | 0 |
| dropout_2 (Dropout) | (None, 16, 64, 128) | 0 |
| conv2d_3 (Conv2D) | (None, 8, 32, 128) | 262272 |
| leaky_re_lu_3 (LeakyReLU) | (None, 8, 32, 128) | 0 |
| dropout_3 (Dropout) | (None, 8, 32, 128) | 0 |
| conv2d_4 (Conv2D) | (None, 4, 16, 128) | 262272 |
| leaky_re_lu_4 (LeakyReLU) | (None, 4, 16, 128) | 0 |
| dropout_4 (Dropout) | (None, 4, 16, 128) | 0 |
| conv2d_5 (Conv2D) | (None, 2, 8, 128) | 262272 |
| leaky_re_lu_5 (LeakyReLU) | (None, 2, 8, 128) | 0 |
| dropout_5 (Dropout) | (None, 2, 8, 128) | 0 |
| flatten (Flatten) | (None, 2048) | 0 |
| dense (Dense) | (None, 1) | 2049 |
| Total params: 1,183,425 | | |
| Trainable params: 1,183,425 | | |
| Non-trainable params: 0 | | |

(2) Architecture of the generator:

It decodes latent space vectors into composite images with dataset features.

First, create a coordinate point input layer and a Dense layer, and then pass them to six Conv2D Transfer layers in sequence (the first five layers have the activation function LeakyReLU, and the last layer has the activation function tanh). The output result is the composite image corresponding to the coordinate point. The model summary is shown in the table below:

Tab.2 model summary of generator

| Layer (type) | Output Shape | Param # |
| --- | --- | --- |
| input_2 (InputLayer) | [(None, 2)] | 0 |
| dense_1 (Dense) | (None, 2048) | 6144 |
| reshape (Reshape) | (None, 2, 8, 128) | 0 |
| conv2d_transpose (Conv2DTran | (None, 4, 16, 128) | 262272 |
| batch_normalization (BatchNo | (None, 4, 16, 128) | 512 |
| leaky_re_lu_6 (LeakyReLU) | (None, 4, 16, 128) | 0 |
| dropout_6 (Dropout) | (None, 4, 16, 128) | 0 |
| conv2d_transpose_1 (Conv2DTr | (None, 8, 32, 128) | 262272 |
| batch_normalization_1 (Batch | (None, 8, 32, 128) | 512 |
| leaky_re_lu_7 (LeakyReLU) | (None, 8, 32, 128) | 0 |
| dropout_7 (Dropout) | (None, 8, 32, 128) | 0 |
| conv2d_transpose_2 (Conv2DTr | (None, 16, 64, 128) | 262272 |
| batch_normalization_2 (Batch | (None, 16, 64, 128) | 512 |
| leaky_re_lu_8 (LeakyReLU) | (None, 16, 64, 128) | 0 |
| dropout_8 (Dropout) | (None, 16, 64, 128) | 0 |
| conv2d_transpose_3 (Conv2DTr | (None, 32, 128, 128) | 262272 |
| batch_normalization_3 (Batch | (None, 32, 128, 128) | 512 |
| leaky_re_lu_9 (LeakyReLU) | (None, 32, 128, 128) | 0 |
| dropout_9 (Dropout) | (None, 32, 128, 128) | 0 |
| conv2d_transpose_4 (Conv2DTr | (None, 64, 256, 64) | 131136 |
| batch_normalization_4 (Batch | (None, 64, 256, 64) | 256 |
| leaky_re_lu_10 (LeakyReLU) | (None, 64, 256, 64) | 0 |
| dropout_10 (Dropout) | (None, 64, 256, 64) | 0 |
| conv2d_transpose_5 (Conv2DTr | (None, 128, 512, 1) | 1025 |
| Total params: 1,189,697 | | |
| Trainable params: 1,188,545 | | |
| Non-trainable params: 1,152 | | |

(3) Model subclassing of Keras to construct WGAN-GP

Just rewrite the __init__(), compile(), metrics(), and train step() methods of the Model class. The other methods inherit.

Import network architecture components and parameters in the __init__() method, provide optimizer and on-screen loss averaging algorithm in the compile() method, and reset the on-screen loss value every epoch in the metrics () method.

The train_step () method: first set the gradient tape, in the gradient tape, utilize the model call () method to get its output, and then calculate the loss; Then calculate the gradient and use the optimizer set by the compile() method to update the weights; Finally, the losses are averaged and then displayed by the fit() method.

Add gradient in the class_ The penaly() method calculates the gradient penalty term (which can also be imported into the class using custom layers).

In the model class, add the gradient_penalty() method (it can also be imported into the model class using a custom layer method).

## 2.3 Training

At the beginning, the critic counts various features of the real image, forcing the generator to generate fake images with some of these features. These fake images sometimes only combine some of the features in the feature set, and this moment the generated images will often appear strong creative.

In order to avoid confusion between fake and real images, the critic further learn and master the matching relationship between different features of real images. Therefore, the generator is further improved to make the composition of fake images almost the same as that of real images, and this moment the generated images will appear mediocre.

So, a moderate model capacity and training epochs can strike a balance between image quality and creativity.

Because there are too many hyperparameters to adjust and hardware constraints (I only have an NVIDIA GTX 1650 graphics card with 4G video memory), it is not possible to fully optimize it. Thus, I can only test parameters while sampling, and generate bridge images from multiple saved generator models.

## 2.4 Exploring new bridge types through latent space sampling

Samples are taken from the latent spaces of multiple generator models according to the set dense spatial grid. Fifty points are taken from each dimension, with a coordinate range of [-10,+10]. Each latent space samples 2500 images (to the 2th power of 50).

Based on the thinking of engineering structure, five technically feasible new bridge types are obtained through manual screening, which were completely different from the dataset (Figure 4).

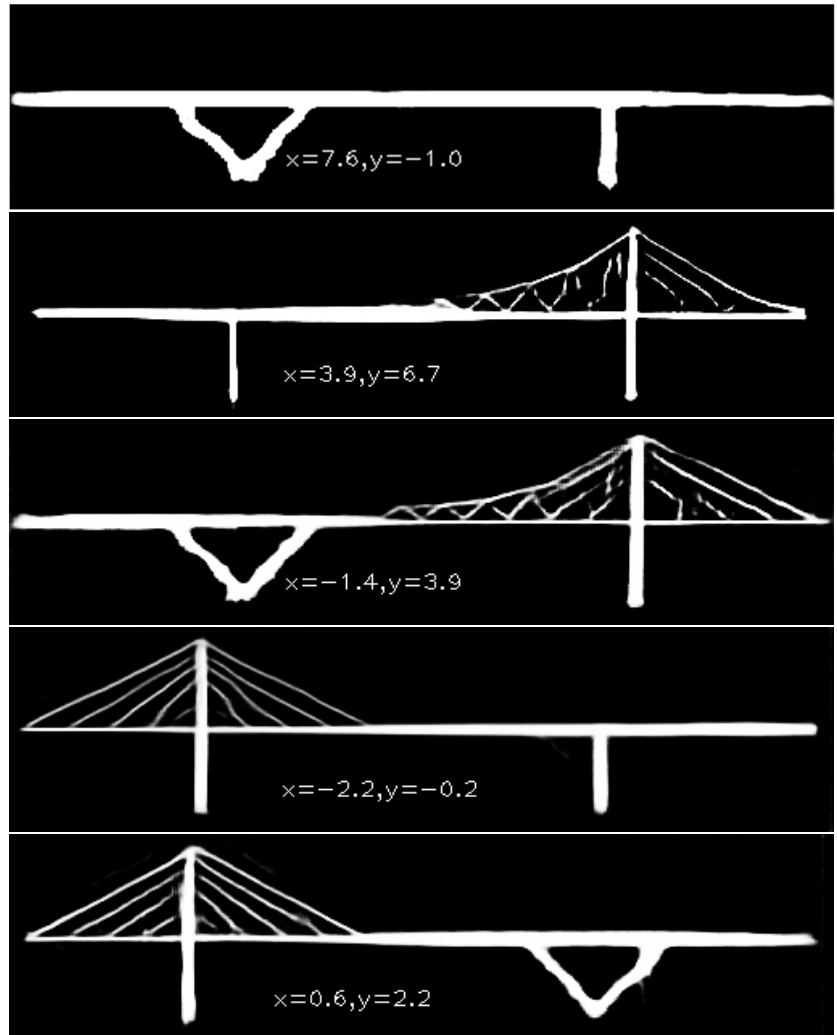

Fig.4 Five new bridge types with feasible technology

The new bridge types here refer to a model that has never appeared in the dataset, but is created by neural network based on algorithms, which represents the model's innovative ability. Some bridge types, such as cable-stayed bridge with single pylon, are very common in reality.

The combination of V-shaped pier rigid structure with suspension bridge with single pylon or cable-stayed bridge with single pylon has not been practiced by any engineer to my knowledge.

## 2.5 Result analysis

The bridge types of the dataset are all symmetric structures, while generative adversarial network can generate asymmetric bridge types, which are not simply superposition, but organic combinations of different structural components. It can be considered that it has similar original human creativity. (Note: The original ability of human bridge design is to combine basic structural components such as beams, columns, cables, and arches according to mechanical principles to form sturdy structures.)

## 2.6 Areas for improvement

(1) After generating tens of thousands of bridge types images through sampling, manual screening based on engineering structural thinking is required. Can program algorithms be used to determine the feasibility of structural mechanics, thus replacing manual image screening?

Even in the discriminator of generative adversarial network, structural mechanics programming algorithm is added to determine whether the structure is valid, so that the latent space of the generator only generates bridge structures that are valid.

(2) There are only eight types of sub bridge types in this dataset, which seriously restricts the innovative ability of generative adversarial network. If we collect all the existing human bridge types, even using three-dimensional bridge types, we believe that low dimensional latent space will generate more new bridge types.

Unfortunately, as an ordinary bridge designer and amateur AI enthusiast, I do not have the resources to conduct in-depth research on the above two points.

## 3  Conclusion

(1) Generative adversarial network is more creative than variational autoencoder. It can organically combine different structural components on the basis of human original bridge types, creating new bridge types. It has a certain degree of human original ability, which can open up imagination space and provide inspiration to humans.

(2) At present, generative artificial intelligence cannot replace the work of human structural engineers, but can only serve as a virtual assistant. After all, structural design is a highly advanced intellectual behavior, and even humans need years of rigorous training to become qualified structural engineers.

Human beings are actually biological robots nurtured by the natural environment. Since nature can generate intelligent humans, as long as its principles are found, humans can also imitate nature and create artificial intelligence. The pursuit of artificial intelligence is to make machines think and act like humans, completely replacing human engineers in the field of engineering. Through continuous research, it is expected that in the near future, "Strong AI" will be achieved. Machines will think and reason like humans, have independent consciousness, and through learning knowledge and engineering practice, can also become qualified engineers for engineering design and construction.

(3) The results of this exploration can be replicated in other design industries, such as industrial design, architectural design, landscape design, etc.